\documentclass[sigconf]{acmart}
\AtBeginDocument{}
\usepackage{amsmath}
\usepackage{graphicx}
\usepackage{multirow}
\usepackage{array}
\usepackage[ruled,vlined,linesnumbered]{algorithm2e}

\setcopyright{none}
\renewcommand\footnotetextcopyrightpermission[1]{}
\acmConference[]{}{}{}
\acmYear{2027}

\AtBeginDocument{%
  \pagestyle{fancy}%
  \fancyhf{}
  \fancyfoot[C]{\thepage}
}

\begin{document}

\title{From Intent to Improvement: Agentic Dataset Search for Better Model Training}
\author{Qiaosheng Chen}
\authornote{These authors contributed equally to this work.}
\orcid{0009-0002-0610-7725}
\affiliation{\institution{Nanjing University}\department{State Key Laboratory for Novel Software Technology}\city{Nanjing}\state{Jiangsu}\country{China}}
\email{qschen@smail.nju.edu.cn}
\author{Kaijia Huang}
\authornotemark[1]
\orcid{0009-0004-0359-7946}
\affiliation{\institution{Nanjing University}\department{State Key Laboratory for Novel Software Technology}\city{Nanjing}\state{Jiangsu}\country{China}}
\email{kaijiahuang@smail.nju.edu.cn}
\author{Fanqi Zhou}
\affiliation{\institution{Nanjing University}\department{State Key Laboratory for Novel Software Technology}\city{Nanjing}\state{Jiangsu}\country{China}}
\email{fqzhou@smail.nju.edu.cn}
\author{Gong Cheng}
\authornote{Corresponding author.}
\orcid{0000-0003-3539-7776}
\affiliation{\institution{Nanjing University}\department{State Key Laboratory for Novel Software Technology}\city{Nanjing}\state{Jiangsu}\country{China}}
\email{gcheng@nju.edu.cn}
\renewcommand{\shortauthors}{Chen et al.}

\begin{abstract}
Finding training data in million-scale machine learning repositories is a Web search problem where text relevance is not enough. A query couples a semantic training goal with constraints such as task, license, and training stage, and the metadata that would verify them are often missing. Text retrieval ranks dataset cards by similarity but cannot enforce hard constraints or follow the relations that reach a sparsely described dataset. Free-form tool-calling agents are more flexible, but the same model that picks a dataset also declares it a match, and nothing checks that the query's constraints are actually met. Neither approach establishes that the data it returns will improve model training. To address it, we propose AgenticDS, a contract-governed agentic framework in which every language model decision instantiates a typed object before deterministic code consumes it. The framework compiles a query into a typed search plan, executes it through admissible deterministic operators over hybrid retrieval and bounded graph expansion, and verifies the provenance-grounded evidence before accepting a ranking or emitting a typed diagnosis that permits only local repairs. On HuDS-Bench, a new benchmark of 200 queries with 24,908 graded judgments pooled over 15 systems, AgenticDS improves NDCG@10 by 25.7\% over the strongest dense--sparse hybrid and by 18.9\% over a commercial LLM agent. A controlled search-to-train study further shows that the datasets it retrieves give the best downstream accuracy in mathematics, code, and medicine. Code and data are publicly available at \url{https://github.com/nju-websoft/AgenticDS}.

\end{abstract}


\maketitle

\section{Introduction}

Open machine learning (ML) platforms, exemplified by Hugging Face,\footnote{\url{https://huggingface.co/}} now host millions of interconnected datasets, models, papers, and curated collections. For a researcher who wants to improve a model, dataset acquisition is already a Web search problem: an underspecified training need must be turned into search operations over inconsistent metadata, relations across heterogeneous resources resolved, and candidates judged by whether they will change the behavior of the model being trained~\citep{chapman2020survey,benjelloun2020google,li2025survey,brickley2019google}. Keyword and embedding-based retrieval help when dataset cards are complete, but they fail on hard structured constraints and ignore evidence implicit in training provenance, task tags, paper citations, or community collections~\citep{chen2024snippets,chen2024dunks,zhou2025mds,chen2025benchmarking,lin2024relationships}.

Natural language dataset acquisition poses three coupled \textbf{technical challenges}. (1)~A query mixes a semantic training goal with typed constraints such as task, modality, license, and training stage. Existing dataset search systems either flatten the query into one embedding~\citep{chen2023dense,zhang2025qwen3} or treat metadata fields independently~\citep{brickley2019google,chen2023integrated}, so \emph{constraint violations stay invisible}. Handing the tools to an LLM restores flexibility but adds no verification, so the same model that selects a dataset also declares it a match~\citep{zhang2024agenticir,jin2025searchr1}. (2)~Evidence is heterogeneous. Text retrieval is precise when dataset cards are well written~\citep{chen2024snippets,zhou2025mds}, while knowledge graph (KG) relations recover resources whose descriptions are sparse~\citep{luo2023empirical,chen2025benchmarking,lin2024relationships}. Structural connectivity does not imply semantic fit, however, so a naively added \emph{graph channel can hurt relevance} when nothing suppresses the noise it introduces~\citep{luo2023empirical}. (3)~Retrieval relevance does not entail training utility. Standard benchmarks score ranking against graded relevance judgments~\citep{lin2022acordar,chen2024acordar2} but \emph{never measure whether the selected data improve a downstream model}, so linking search quality to data-centric recursive self-improvement (RSI), the loop in which evaluation results drive changes to the training data~\citep{zou2026datacentric}, requires a protocol that measures that change directly.

To address these challenges, we present \textbf{AgenticDS}, a contract-governed agentic framework for Web dataset acquisition. A \emph{Planner} compiles the query into a typed search plan of constraints, allowed KG relations, and acceptance criteria. An \emph{Engineer} executes only admissible deterministic operators over hybrid retrieval, bounded KG expansion, and provenance-preserving pre-ranking. A \emph{Critic} scores a fixed-format evidence package and either accepts the result or emits a typed diagnosis permitting only local plan repairs. The contracts stop an LLM from silently redefining retrieval semantics during tool use, while the evidence trace keeps text, metadata, content, and KG support separately inspectable.

\begin{figure*}[t]
\centering
\includegraphics[width=\textwidth]{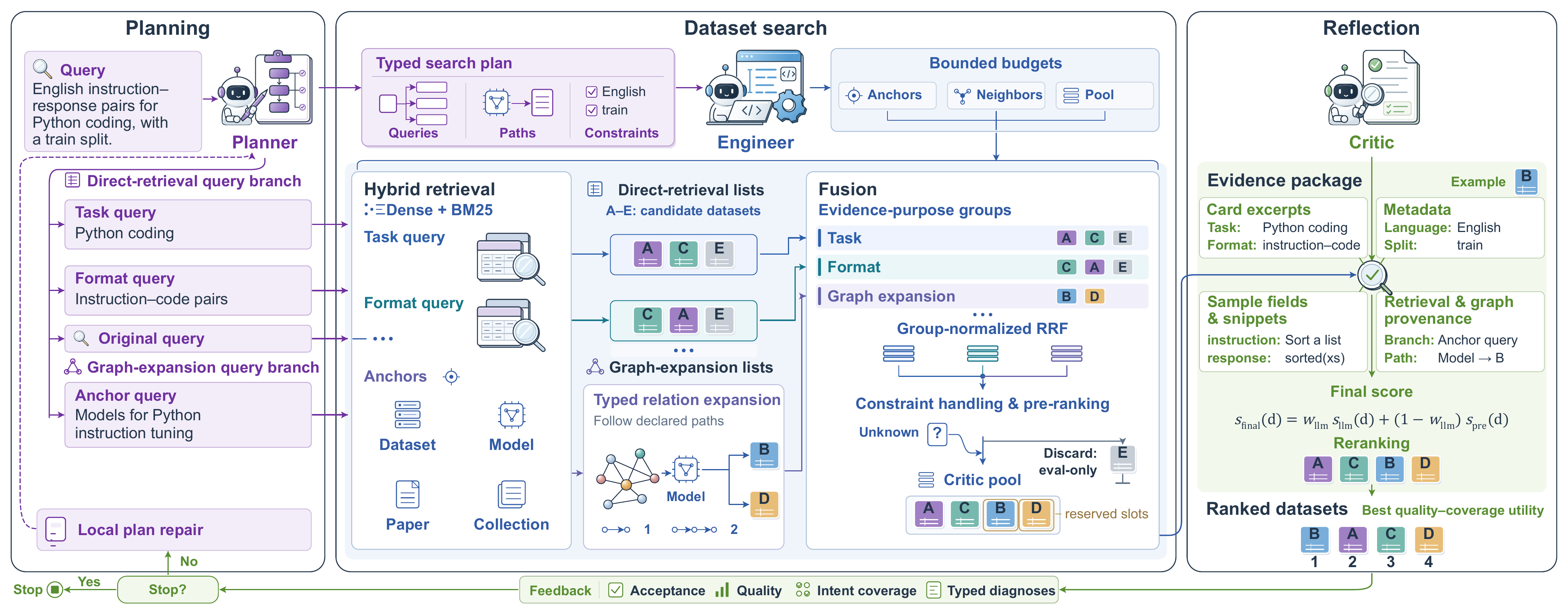}
\caption{Overview of AgenticDS.}
\Description{A Planner constructs complementary query branches. An Engineer allocates bounded resources and deterministically executes typed retrieval, directional expansion, branch fusion, and Critic-pool construction. The Critic independently scores evidence and diagnoses local repairs.}
\label{fig:overview}
\end{figure*}

To evaluate this design, we build \textbf{HuDS-Bench}, a KG-grounded benchmark of 200 natural language dataset search queries with 24{,}908 graded judgments pooled over 15 systems. AgenticDS outranks every system in the pool, and removing the Critic or the Planner costs 0.149 and 0.086 NDCG@10. We then test whether better retrieval translates into better training, feeding each system's top-5 datasets into corresponding supervised fine-tuning (SFT) of three base models across mathematics, code, and medicine. AgenticDS gives the best downstream accuracy in all three domains, with the largest gain on the weakest base model, supporting the view that external acquisition complements within-corpus data-centric RSI. Ablation then isolates where relational evidence pays: graph expansion whose candidates pass straight into the ranking lowers accuracy on every text backbone, whereas the same expansion behind a provenance-aware Critic improves it, so expansion coupled to verification rather than the graph itself is the mechanism.

In summary, our contributions are threefold:
\begin{itemize}
  \item We formulate natural language dataset acquisition on the Web as contract-governed agentic retrieval, and present AgenticDS, whose typed plans, evidence records, and diagnoses separate semantic reasoning from deterministic execution.
  \item We show that relational evidence is conditional rather than additive, and contribute the per-candidate provenance representation that lets a verifier turn graph expansion from a liability into a gain.
  \item We release HuDS-Bench and a search-to-train protocol that measures whether retrieved datasets improve a model, and use them to compare 15 retrieval systems and to fine-tune three base models across mathematics, code, and medicine.
\end{itemize}

\section{Related Work}

\subsection{Dataset Search on the Web}
Dataset search differs from document retrieval because relevance depends on structured constraints, data content, and intended use beyond text~\citep{chapman2020survey}. Beyond Google Dataset Search's Web metadata index~\citep{brickley2019google}, later work targets dense representations~\citep{chen2023dense,luo2023empirical}, query-focused snippets~\citep{chen2024snippets,chen2024dunks,zhou2025mds}, and integrated public portals~\citep{chen2023integrated}. Closest to our setting, DataFinder recommends scientific datasets from a natural language description of a research idea~\citep{viswanathan2023datafinder}. It maps one free-text need to a flat bi-encoder ranking over a fixed corpus, without typed constraints, relational evidence, or any check that the recommended data help training. \emph{These are the three axes this paper adds}. Benchmarks such as ACORDAR evaluate public-data retrieval~\citep{lin2022acordar,chen2024acordar2} but do not model natural language data needs over modern ML hubs. Relational evidence has been explored through general knowledge bases~\citep{luo2023empirical} and taxonomies of dataset relationships~\citep{lin2024relationships}. HuggingKG links datasets with models, tasks, papers, spaces, and collections~\citep{chen2025benchmarking}. \emph{These lines supply the primitives and relational infrastructure AgenticDS composes under a typed contract}. We adopt the HuggingKG snapshot as auxiliary infrastructure rather than as a contribution of this paper, using it as one candidate source alongside dense and sparse retrieval, and \emph{design HuDS-Bench so that natural language data needs become directly measurable}.

\subsection{Agentic Information Retrieval}
LLM agents combine planning, tool use, and reflection~\citep{yao2023react}, building on retrieval-augmented generation~\citep{lewis2020rag} and LLM rankers~\citep{sun2023chatgpt}. Zhang et al.~\citep{zhang2024agenticir} formalize ``agentic information retrieval'' as agents that plan, execute, and refine multi-step search strategies. Search-R1~\citep{jin2025searchr1} learns when to call a search engine during reasoning, and Search-o1~\citep{li2025searcho1} embeds agentic Web search into extended reasoning chains. A study of 14 million real queries shows that agentic intents and trajectories differ markedly from traditional Web search~\citep{ning2026agenticsearch}. Most such systems target open-domain question answering and give the LLM direct control of tools, which entangles query interpretation with index operations and makes errors and reruns hard to compare. Commercial coding agents prompted to search Hugging Face behave the same way. \emph{AgenticDS instead keeps semantic reasoning inside typed contracts and leaves indexing, traversal, filtering, and scoring deterministic}, trading open-ended tool flexibility for reproducibility, provenance-transparent ranking, and bounded token cost.

\subsection{Data-Centric Recursive Self-Improvement}
Data-centric RSI systems filter, select, rewrite, or synthesize samples from a known corpus. DataEvolver learns preparation pipelines from raw data and seeds through operator- and pipeline-level feedback~\citep{deng2026dataevolver}. DataFoundry evolves a data-construction runtime and synthesizes instruction--response pairs from supplied documents~\citep{yang2026datafoundry}. The broader literature studies evaluation-driven updates to data, retrieval state, memory, or policies~\citep{zou2026datacentric,chen2024spin,yuan2024selfrewarding,wang2024lance}. \emph{All presuppose a corpus already in hand and optimize data within it, while our work targets the preceding step}. We acquire external datasets from an open Web ecosystem given a user query, and verify utility through controlled fine-tuning. That search-to-train primitive is the ingredient these systems assume, positioning AgenticDS as a complementary upstream component of future data-centric loops rather than a competing in-corpus optimizer.

\section{AgenticDS: Framework and Mechanism}
\label{sec:method}

\subsection{Task and Design Principles}
\label{sec:method_task}
Let $q$ be a query containing a natural language description of a user's training data need, and $\mathcal D$ a dataset collection in which each $d\in\mathcal D$ carries heterogeneous and often incomplete metadata. We seek a ranked list $\mathcal L(q)$ of up to $K$ datasets that meet the goal and hard constraints of $q$ and rank highly on its soft preferences. Our implementation uses a heterogeneous graph $\mathcal G$ from HuggingKG (Appendix~\ref{app:resource}), whose nodes include datasets, models, papers, and collections, with relations such as model--training-dataset and collection membership. Missing metadata or graph links indicate unknown support, not irrelevance.

AgenticDS follows three principles. \emph{Contracts before actions} makes typed plans delimit the admissible operators. \emph{Provenance before fusion} keeps evidence sources separately inspectable. \emph{Bounded autonomy} lets budgets and diagnoses constrain search and local plan repairs. These principles bound what the LLM may redefine without discarding the semantic flexibility needed to interpret complex queries. Figure~\ref{fig:overview} gives an overview of the pipeline, and Algorithm~\ref{alg:agenticds} formalizes the search loop.

\begin{algorithm}[t]
\small
\caption{AgenticDS search loop.}
\label{alg:agenticds}
\KwIn{query $q$, corpus $\mathcal D$, graph $\mathcal G$, result count $K$, round limit $T$}
$\mathcal B,\mathcal Q,R_{\rm best},\mathbf P_0,\mathcal F_0\leftarrow\varnothing$\nllabel{ln:initialize}\;
\For{$t=1$ \KwTo $T$}{
 $\mathbf P_t\leftarrow\mathrm{Planner.Plan}(q,K,\mathcal G,\mathbf P_{t-1},\mathcal F_{t-1},\mathcal Q)$\nllabel{ln:planner}\;
 \lIf{$\neg\mathrm{Planner.HasNewQuery}(\mathbf P_t,\mathcal Q)$\nllabel{ln:nonew}}{\textbf{break}}
 $\mathbf S_t\leftarrow\mathrm{Engineer.Compile}(\mathbf P_t)$\nllabel{ln:engineer}\;
 $(\{L_j^{\rm direct}\},A_t)\leftarrow\mathrm{Engineer.HybridRetrieve}(\mathbf S_t,\mathcal D,\mathcal G)$\nllabel{ln:hybrid}\;
 $\{L_j^{\rm graph}\}\leftarrow\mathrm{Engineer.TypedKGExpand}(A_t,\mathcal G,\mathbf S_t)$\nllabel{ln:expand}\;
 $\mathcal B\leftarrow\mathrm{Engineer.UpdateBranchLists}(\mathcal B,\{L_j^{\rm direct}\},\{L_j^{\rm graph}\})$\nllabel{ln:branchlists}\;
 $\mathcal Q\leftarrow\mathcal Q\cup\mathrm{Signatures}(\mathcal H_t)$\nllabel{ln:queries}\;
 $s_{\rm RRF}(\cdot)\leftarrow\mathrm{Engineer.GroupRRF}(\mathcal B,\mathbf P_t)$\nllabel{ln:fusion}\;
 $C_t\leftarrow\mathrm{Engineer.ConstraintPrerank}(\mathcal B,s_{\rm RRF},\mathbf P_t,\mathbf S_t)$\nllabel{ln:prerank}\;
 $(R_t,\mathcal F_t)\leftarrow\mathrm{Critic.Review}(q,\mathbf P_t,C_t)$\nllabel{ln:critic}\;
 $R_{\rm best}\leftarrow R_{\arg\max_{1\le s\le t}U(\mathcal F_s)}$\nllabel{ln:best}\;
 \lIf{$\mathrm{Stop}(\mathcal F_{1:t})$\nllabel{ln:stop}}{\textbf{break}}
}
\Return $R_{\rm best}[1:K]$\nllabel{ln:return}\;
\end{algorithm}

\subsection{Planner: Compiling Intent into a Typed Contract}
The Planner produces a typed search plan (line~\ref{ln:planner}) from $q$, $K$, and the allowed path types of $\mathcal G$. On later rounds it also reads the previous plan $\mathbf P_{t-1}$, feedback $\mathcal F_{t-1}$, and executed-query set $\mathcal Q$, all initialized empty in line~\ref{ln:initialize}. In local repair the Planner preserves the original requirements while adding new queries that address diagnosed gaps. The Planner ends the search once it has no new query, that is, no query-branch signature outside $\mathcal Q$ (line~\ref{ln:nonew}). The contract is
\begin{equation}
\mathbf P_t=(\mathcal H_t,\mathcal R_t,\mathcal C_{\rm hard},\mathcal C_{\rm soft},I_q,\mathcal N_t,\mathcal T_{\rm qual}).
\label{eq:plan}
\end{equation}
In Eq.~\eqref{eq:plan}, $\mathcal H_t$ contains several query branches, each grounded in $q$ and associated with a specific evidence purpose. The Planner marks a branch \emph{direct-retrieval} when its candidates come from the dataset index itself, and keeps one such branch at the original user query, preserving joint semantics that decomposition may separate. It marks a branch \emph{graph-expansion} when its candidates must instead be reached from entity anchors over declared graph paths. An \emph{anchor} starts a graph path and may be a dataset, model, paper, or collection. In $\mathcal R_t$ the Planner declares the one- or two-hop paths to datasets in $\mathcal G$. It records requirements in $\mathcal C_{\rm hard}$, preferences in $\mathcal C_{\rm soft}$, and the intents a candidate should jointly support in $I_q$. In $\mathcal N_t$ it reserves bounded pool slots for selected query branches, because broad text matches could otherwise crowd out candidates with specific provenance or evidence before verification. In $\mathcal T_{\rm qual}$ it sets quality, intent-coverage, and minimum-count thresholds.

As illustrated in Figure~\ref{fig:overview}, the Planner emits a Task query branch targeting Python coding, a Format query branch targeting instruction--code pairs, and a graph-expansion query branch anchored on models for Python instruction tuning, alongside the retained original query.

\subsection{Engineer: Hybrid Retrieval and Typed Relation Expansion}
The Engineer compiles $\mathbf P_t$ into an action protocol $\mathbf S_t$ (line~\ref{ln:engineer}): predefined operators with bounded anchor, neighbor, and pool budgets. Within these budgets, the Engineer adaptively allocates retrieval effort according to $\mathbf P_t$, which indicates where useful evidence is likely to lie. It then applies hybrid retrieval to collect candidate datasets for direct-retrieval query branches and anchors for graph-expansion query branches (line~\ref{ln:hybrid}). For a retrieved entity (i.e., a candidate dataset or an anchor) $e$, its dense score is
\begin{equation}
s_{\rm dense}(e)=\eta s_{\rm desc}(e)+(1-\eta)s_{\rm meta}(e),
\label{eq:dense}
\end{equation}
where $s_{\rm desc}$ and $s_{\rm meta}$ are cosine similarities between the branch query and the entity's description and metadata embeddings. Equation~\eqref{eq:dense} balances semantic descriptions and structured attributes through configurable $\eta\in[0,1]$. Dense and BM25 ranks are combined by reciprocal rank fusion (RRF)~\citep{cormack2009rrf}:
\begin{equation}
s_{\rm hybrid}(e)=\alpha(k+r_{\rm dense}(e))^{-1}+(1-\alpha)(k+r_{\rm BM25}(e))^{-1}.
\label{eq:rrf}
\end{equation}
In Eq.~\eqref{eq:rrf}, $r_{\rm dense}$ and $r_{\rm BM25}$ denote the ranks of $e$ in the dense and BM25 retrieval lists, respectively; $\alpha\in[0,1]$ controls their relative contributions, while $k>0$ smooths rank differences. The outputs are direct-retrieval lists $L_j^{\rm direct}$, indexed by query branch $j$, and anchors $A_t$ for graph-expansion query branches.

The Engineer then performs typed KG expansion from $A_t$, following only the paths declared in $\mathcal R_t$, to produce graph-expansion lists $L_j^{\rm graph}$ (line~\ref{ln:expand}). Let $\mathcal P_d$ be the permitted paths from anchors in $A_t$ to dataset $d$, each written $p=(a=v_0,\ldots,v_l=d)$ with edge types $r_h$ for $1\le h\le l$. The relation score of dataset $d$ is
\begin{equation}
\begin{aligned}
s_{\rm rel}(d)={}&\sum\nolimits_{p\in\mathcal P_d}s_{\rm hybrid}(a)(\rho+(1-\rho)c(a))\\
&\cdot\gamma^{l-1}\prod\nolimits_{h=1}^{l}[\log_2(2+\deg_{r_h}(v_{h-1}))]^{-1}.
\end{aligned}
\label{eq:kg}
\end{equation}
Equation~\eqref{eq:kg} aggregates support from all permitted anchor-to-dataset paths. Here $s_{\rm hybrid}(a)$ is given by Eq.~\eqref{eq:rrf}, $c(a)\in[0,1]$ measures its name/identifier match to the query, and $\deg_r(v)$ counts $v$'s neighbors under relation $r$. Configurable $\rho\in[0,1]$ and $\gamma\in(0,1]$ set the confidence floor and path decay. This lets AgenticDS reach a sparsely described dataset through a supported entity anchor. The degree-based penalties reduce support from widely connected entities, while path decay discounts indirect paths. In Figure~\ref{fig:overview}, the Engineer retrieves a relevant model, follows its training-data links to candidate datasets B and D, and orders them by relation score.

\begin{table*}[t]
\centering
\caption{Composition of HuDS-Bench.}
\label{tab:benchmark_main}
\begin{tabular}{lrlp{0.55\textwidth}}
\toprule
Query type & Count & Primary grounding & Typical requirement \\
\midrule
General & 90 & Metadata + task tags & Discover data satisfying task, language, modality, or scale constraints \\
Fine-tuning & 40 & Metadata + training stage & Find instruction, dialogue, preference, or domain-adaptation data \\
Paper reference & 40 & Paper/model relations & Locate data associated with a specific paper or training lineage \\
Entity neighbor & 30 & Dataset/collection relations & Find datasets reachable through multi-hop collection, model, or task paths \\
\bottomrule
\end{tabular}
\end{table*}

\subsection{Constraint Handling and Provenance-Preserving Pre-Ranking}
The Engineer updates the branch lists held in the cross-round archive $\mathcal B$ with $\{L_j^{\rm direct}\}$ and $\{L_j^{\rm graph}\}$, replacing previous lists with the same query-branch identifier (line~\ref{ln:branchlists}). It also adds the signatures of the current query branches to the query history $\mathcal Q$ (line~\ref{ln:queries}). The Engineer then fuses the archive into a single ranking by group-normalized RRF (line~\ref{ln:fusion}). Each list $L_j$ belongs to a group $g(j)$ defined by evidence purpose, such as Task, Format, or Graph expansion. For nonempty lists, weights are $\widetilde w_j=w_j(\sum_{\ell:g(\ell)=g(j)}w_\ell)^{-1}$, where $w_j\ge0$ weights query branch $j$ and group totals are positive. With $Z=\sum_j\widetilde w_j$ and list rank $r_j(d)$,
\begin{equation}
s_{\rm RRF}(d)=(k+1)Z^{-1}\sum\nolimits_j\widetilde w_j\mathbf1[d\in L_j](k+r_j(d))^{-1}.
\label{eq:constraint}
\end{equation}
Equation~\eqref{eq:constraint} bounds the fusion score of dataset $d$ within $[0,1]$ and prevents additional queries from disproportionately increasing the influence of any single group.

The Engineer then pre-ranks the fused candidates, discarding those with explicit hard-constraint violations (line~\ref{ln:prerank}). Missing fields remain unknown, so potentially relevant datasets are not rejected solely for incomplete metadata. Satisfactory candidates precede unknown ones, and within each of those two classes Eq.~\eqref{eq:prerank} adjusts retrieval support for soft preferences:
\begin{equation}
s_{\rm pre}(d)=\beta s_{\rm RRF}(d)+(1-\beta)U_{\rm pref}(d),
\label{eq:prerank}
\end{equation}
where configurable $\beta\in[0,1]$ balances relevance and preferences, and $U_{\rm pref}\in[0,1]$ is a weighted mean of preference utilities. For download or like count $x\ge0$, the utility is $\min\{1,\log(1+x)/\log(1+\tau)\}$, with configurable saturation threshold $\tau>0$; minimization uses its complement. Without preferences, we set $s_{\rm pre}=s_{\rm RRF}$. From this ordering and the bounded reservations $\mathcal N_t$ the Engineer selects the Critic pool $C_t$. In Figure~\ref{fig:overview}, for the train-split requirement above, dataset E explicitly marked evaluation-only is discarded, while an unreported split remains unknown. Candidates retain their query-branch ranks, fusion contributions, constraint status, anchors, and paths.
Retaining this record rather than collapsing it into a single score makes the ranking auditable, because a user can tell whether a candidate is supported by its dataset card or by a relation path.

\subsection{Critic: Evidence Compression, Reranking, and Diagnosis}
The Critic reviews the pool $C_t$ against $q$ and $\mathbf P_t$ (line~\ref{ln:critic}). Each candidate has an evidence package with four parts: concise card excerpts, normalized metadata, sample fields and instruction--response snippets when available, and retrieval and graph provenance. These sources support judgments of semantic fit, constraint satisfaction, and relational support from the graph. Conditioned on $q$ and $\mathbf P_t$, the Critic judges how strongly each evidence package supports joint satisfaction of the query requirements and returns a semantic score $s_{\rm llm}(d)\in[0,1]$, combined with its deterministic pre-score $s_{\rm pre}(d)$:
\begin{equation}
s_{\rm final}(d)=w_{\rm llm}s_{\rm llm}(d)+(1-w_{\rm llm})s_{\rm pre}(d).
\label{eq:final}
\end{equation}
The configurable $w_{\rm llm}\in[0,1]$ balances semantic judgment and retrieval support in Eq.~\eqref{eq:final}, allowing content evidence to refine the retrieval ranking while retaining its support. Sorting candidates by $s_{\rm final}(d)$ yields the current ranking $R_t$. In Figure~\ref{fig:overview} this lifts the graph-reached dataset B to the top of the pool.

In $\mathcal F_t$ the Critic records acceptance, ranking quality $u_t$, intent coverage $v_t$, and typed diagnoses. Quality is the mean of final scores over the top $\lceil K/4\rceil$ datasets in $R_t$ (all if fewer), emphasizing the first results users inspect. Intent coverage is the maximum fraction of intents in $I_q$ supported by at least one dataset among the top $\lceil K/2\rceil$ results. Thus, a Python corpus and an unrelated instruction corpus cannot jointly establish instruction--response pairs for Python coding. Acceptance requires that ranking quality and intent coverage meet their respective thresholds in $\mathcal T_{\rm qual}$, that the minimum result count is satisfied, and that no critical diagnosis remains unresolved. Otherwise the Critic emits typed diagnoses that guide next-round local plan repair.

AgenticDS retains the round of maximal utility $U(\mathcal F_s)=\lambda u_s+(1-\lambda)v_s$ over $s\le t$ (line~\ref{ln:best}), where configurable $\lambda\in[0,1]$ balances quality against joint intent coverage. Given the feedback history $\mathcal F_{1:t}$, AgenticDS stops on deterministic criteria, namely acceptance, two consecutive rounds without utility improvement, or a critical diagnosis recurring in consecutive rounds (line~\ref{ln:stop}). Finally, AgenticDS returns the first $K$ datasets of $R_{\rm best}$ (line~\ref{ln:return}), which is the ranked list $\mathcal L(q)$ in Section~\ref{sec:method_task}.

Appendix~\ref{app:method} lists all contracts, score weights, and prompts. Because the Planner fixes the query branches and the Engineer may run only the declared operators, a module can be disabled while everything downstream of it behaves as before, which is what makes component-level ablation well defined.

\section{Evaluation Methodology}
\label{sec:evaluation}

We evaluate two research questions. \textbf{RQ1}: Can AgenticDS retrieve datasets that satisfy complex natural language training needs on a modern ML hub? \textbf{RQ2}: Can the retrieved datasets improve models when used as training interventions? RQ1 uses a dataset search benchmark and ranking metrics. RQ2 holds training budgets and optimization settings constant so that differences arise from dataset acquisition rather than training recipe.

\subsection{HuDS-Bench}
HuDS-Bench targets realistic Web dataset search over HuggingFace and consists of 200 English queries of four types, two \emph{semantic} (general and fine-tuning) and two \emph{relational} (paper-reference and entity-neighbor). Its construction is KG-grounded. An extractor derives structured conditions from a gold dataset's metadata and, for relational types, from HuggingKG~\citep{chen2025benchmarking} anchors. An LLM, GLM-5.2, writes a concise user need from a controlled subset of those conditions, seeing the anchor and target relation but not the gold dataset's identity or card text for the two relational types. Rule-based checks and manual verification then reject queries that expose the gold dataset's HuggingFace identifier, copy long card spans, omit intent, or collapse to an exact-name lookup.

The four types stress complementary evidence channels (Table~\ref{tab:benchmark_main}). The semantic types emphasize semantic goals plus structured constraints that a dataset card can usually answer. The relational types instead anchor on a paper, model, or collection, and 29 of the 30 entity-neighbor queries are strictly multi-hop. KG relations therefore enter benchmark construction, so the benchmark can produce needs whose correct answer is unreachable by lexical matching alone.

\subsection{Pooling and Relevance Annotation}
Candidate pools merge the top-20 results of all 15 evaluated systems, with the gold dataset injected only when no system retrieves it. The pool holds 24{,}908 query--candidate pairs. Rather than pass raw dataset cards to an annotator, we compile each query into a structured rubric and assemble a per-candidate evidence package, from which the LLM annotator, DeepSeek-V4-Pro-0813, resolves each constraint to a three-valued status that we deterministically project to a graded 0/1/2 label for the candidate. A second, independent pass by the same model re-checks every disputed item, and residual disagreements go to human adjudication. Against a 2{,}490-pair human-annotated sample the pipeline reaches linearly weighted Cohen's $\kappa=0.754$, substantial agreement on the conventional scale~\citep{landis1977kappa}. Appendix~\ref{app:benchmark} gives the rubric, the label distribution by query type, and the annotation procedure.

\subsection{Baselines, Backbone, and Metrics}
\label{sec:evaluation-baselines-backbone-metrics}
We compare AgenticDS against four strong baselines spanning the Web-agent and traditional retrieval families. Two \emph{LLM agents} are implemented with Claude Code and Codex harnesses, both driven by GLM-5.2~\citep{glm5} and given the same HuggingFace Hub tools. \emph{Qwen3-Embedding-0.6B + BM25} is a dense--sparse hybrid fused by RRF~\citep{zhang2025qwen3,cormack2009rrf}, and \emph{monoT5 (OOD) + BM25} is a classic retrieval-then-rerank pipeline~\citep{nogueira2020document}. We mark monoT5 \emph{(OOD)} because it is applied zero-shot, no in-domain training signal existing for this task. Nine further systems span \emph{platform search, sparse, dense, and hybrid retrieval}. Appendix~\ref{app:implementation} lists every system with its configuration and gives the agent tool interface and execution protocol.

AgenticDS runs with GLM-5.2 as the agent backbone, Qwen3-Embedding-0.6B as its embedding model, and $T=3$ iterations unless stated. We also report a second variant that swaps only the embedding model for bge-m3~\citep{chen2024m3}. Because the same model family generated the HuDS-Bench queries, a shared prior over phrasing could flatter the system. The backbone sweep bounds that concern, since substituting Qwen3.5-35B-A3B from a different family still places AgenticDS above every baseline (Appendix~\ref{app:extended}). All hyperparameters are fixed across reported runs, tuned beforehand on a pilot set disjoint from HuDS-Bench.

For RQ1, retrieval quality is measured by NDCG and MAP at cutoffs 5/10/20. Statistical significance is assessed by two-sided paired permutation tests over the 200 per-query scores with Holm correction within each family of comparisons, and confidence intervals by paired bootstrap resampling stratified by query type. We fix AgenticDS (Qwen3-Emb) as the reference and correct over the 14 comparisons against the other systems. Appendix~\ref{app:significance} gives the resampling configuration and every test outcome.

\begin{table}[t]
\centering
\caption{Retrieval quality on HuDS-Bench. Bold marks the best system and underline the second best. $^{\dag}$ marks a cell where AgenticDS (Qwen3-Emb) is significantly higher than that system ($p<0.05$).}
\label{tab:retrieval}
\small
\setlength{\tabcolsep}{3.1pt}
\begin{tabular}{lccc@{\hskip 7pt}ccc}
\toprule
& \multicolumn{3}{c}{NDCG} & \multicolumn{3}{c}{MAP} \\
\cmidrule(lr){2-4}\cmidrule(lr){5-7}
Method & @5 & @10 & @20 & @5 & @10 & @20 \\
\midrule
\multicolumn{7}{l}{\textit{Platform search}}\\
GoogleDS & .0301$^{\dag}$ & .0217$^{\dag}$ & .0165$^{\dag}$ & .0262$^{\dag}$ & .0146$^{\dag}$ & .0091$^{\dag}$ \\
HFFullText & .2786$^{\dag}$ & .2438$^{\dag}$ & .2100$^{\dag}$ & .2989$^{\dag}$ & .2301$^{\dag}$ & .1661$^{\dag}$ \\
\midrule
\multicolumn{7}{l}{\textit{Sparse}}\\
BM25 & .4719$^{\dag}$ & .4717$^{\dag}$ & .4797$^{\dag}$ & .5912$^{\dag}$ & .5575$^{\dag}$ & .5252$^{\dag}$ \\
BM25+RM3 & .4772$^{\dag}$ & .4703$^{\dag}$ & .4732$^{\dag}$ & .5994$^{\dag}$ & .5482$^{\dag}$ & .5047$^{\dag}$ \\
\midrule
\multicolumn{7}{l}{\textit{Dense}}\\
bge-m3 & .3102$^{\dag}$ & .3204$^{\dag}$ & .3309$^{\dag}$ & .4751$^{\dag}$ & .4475$^{\dag}$ & .4084$^{\dag}$ \\
Qwen3-Emb & .4728$^{\dag}$ & .4699$^{\dag}$ & .4808$^{\dag}$ & .6129$^{\dag}$ & .5830$^{\dag}$ & .5570$^{\dag}$ \\
\midrule
\multicolumn{7}{l}{\textit{Hybrid / rerank}}\\
bge-m3+BM25 & .4039$^{\dag}$ & .4036$^{\dag}$ & .3972$^{\dag}$ & .5507$^{\dag}$ & .5144$^{\dag}$ & .4659$^{\dag}$ \\
MDRBM25 & .4602$^{\dag}$ & .4578$^{\dag}$ & .4698$^{\dag}$ & .5804$^{\dag}$ & .5420$^{\dag}$ & .5107$^{\dag}$ \\
ANCE (OOD)+BM25 & .4617$^{\dag}$ & .4570$^{\dag}$ & .4644$^{\dag}$ & .5913$^{\dag}$ & .5469$^{\dag}$ & .5069$^{\dag}$ \\
Qwen3-Emb+BM25 & .4911$^{\dag}$ & .4893$^{\dag}$ & .4964$^{\dag}$ & .6202$^{\dag}$ & .5919$^{\dag}$ & .5663$^{\dag}$ \\
monoT5 (OOD)+BM25 & .4976$^{\dag}$ & .4923$^{\dag}$ & .5006$^{\dag}$ & .6140$^{\dag}$ & .5795$^{\dag}$ & .5525$^{\dag}$ \\
\midrule
\multicolumn{7}{l}{\textit{LLM agents}}\\
Claude Code & .5328$^{\dag}$ & .5170$^{\dag}$ & .5169$^{\dag}$ & .7009 & .6700 & \underline{.6381} \\
Codex & .5370$^{\dag}$ & .5140$^{\dag}$ & .5082$^{\dag}$ & .7094 & .6764 & .6313 \\
\midrule
\multicolumn{7}{l}{\textit{Ours}}\\
AgenticDS (bge-m3) & \underline{.5692}$^{\dag}$ & \underline{.5518}$^{\dag}$ & \underline{.5424}$^{\dag}$ & \textbf{.7374} & \underline{.6871} & .6311 \\
AgenticDS (Qwen3-Emb) & \textbf{.6297} & \textbf{.6148} & \textbf{.6114} & \underline{.7316} & \textbf{.6973} & \textbf{.6630} \\
\bottomrule
\end{tabular}
\end{table}

\subsection{Utility-Grounded Search-to-Train Protocol}
Ranking does not establish training utility. For RQ2 we treat each search output as a controlled training intervention. For retrieval method $R$ and query $q$, let $E(\mathcal{L}_R(q))$ be the eligible datasets among its top-$K$ results, and let $\mathrm{Sample}(d,b)$ draw at most $b$ samples from dataset $d$. The acquired corpus is
\begin{equation}
\mathcal{A}_R(q)=\bigcup_{d\in E(\mathcal{L}_R(q))}\mathrm{Sample}(d,b),
\label{eq:acq}
\end{equation}
with identical $b=2000$, eligibility rules, and $K=5$ for every method. For base model $f_0$, training algorithm $\mathrm{Train}$, and downstream-task benchmark set $\mathcal{B}_q$ scored by metric $m_B$, acquisition utility is
\begin{equation}
U(R,q)=\tfrac{1}{|\mathcal{B}_q|}\sum_{B\in\mathcal{B}_q}\!\big[m_B(\mathrm{Train}(f_0,\mathcal{A}_R(q)))-m_B(f_0)\big].
\label{eq:utility}
\end{equation}
Only the retrieval method varies. Base model, data budget, trainer, and evaluation suite are fixed.

We instantiate the protocol with three base models (Qwen2.5-7B~\citep{qwen25}, Llama-3.1-8B~\citep{llama3}, and Olmo-3-7B-Instruct~\citep{olmo3}), across three domains (mathematics, code, and medicine) with one realistic query per domain. Learning rate and epoch are selected on a held-out development split, never on the evaluation benchmarks, and we report the test score of the checkpoint that development selection picked. Scores are calculated for a grid of three base models and eight benchmarks: \{GSM8K~\citep{gsm8k}, MATH~\citep{math}, AMC23, OCWCourses~\citep{lewkowycz2022minerva}\} for math, \{HumanEval~\citep{humaneval}, BigCodeBench~\citep{bigcodebench}\} for code, and \{MedQA~\citep{medqa}, MMLU-Medical~\citep{mmlu}\} for medicine, then averaged across base models. The same split, selection procedure, and grid apply to every method, so none is given more selection freedom than another. For statistical significance we block by (domain, base model) and enumerate all $2^9$ sign flips of the nine block differences, Holm-correcting the comparison of AgenticDS with four strong baselines (Section~\ref{sec:evaluation-baselines-backbone-metrics}). Because those blocks share three training corpora and the study has no independent training seeds, we read these tests as a consistency check across blocks rather than as a measure of how much retraining would move the scores. 

Appendix~\ref{app:trainingdetails} gives the benchmark splits, the training grid and search queries, the per-model scores, and the full ablation grid.

\begin{figure}[t]
\centering
\includegraphics[width=\columnwidth]{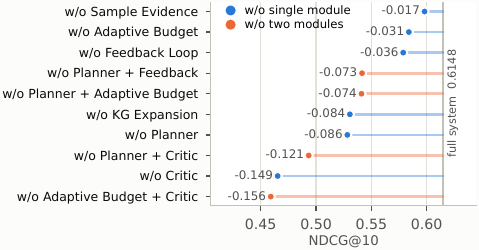}
\caption{Search-pipeline ablation. Each stem is a configuration's distance from the full system (vertical line).}
\Description{Lollipop chart of ten ablated configurations, each a horizontal stem running back to a vertical reference line at the full system's 0.6148 NDCG@10, with single-module removals in blue and two-module removals in orange, ranging from -0.017 for sample evidence to -0.156 for removing both the adaptive budget and the Critic.}
\label{fig:abl_retr}
\end{figure}

\begin{table*}[t]
\centering
\caption{Training utility on eight benchmarks from three domains. Each block averages the three base models. \emph{Avg} is the within-domain mean and \emph{All} the mean of the three domain averages. Bold marks the best system.}
\label{tab:e2e}
\footnotesize
\setlength{\tabcolsep}{5.5pt}
\begin{tabular}{lrrrrrrrrrrrr}
\toprule
& \multicolumn{5}{c}{Math (EM)} & \multicolumn{3}{c}{Code (pass@1)} & \multicolumn{3}{c}{Medicine (acc.)} & \\
\cmidrule(lr){2-6}\cmidrule(lr){7-9}\cmidrule(lr){10-12}
Acquisition method & GSM8K & MATH & AMC23 & OCWCourses & \textit{Avg} & HumanEval & BigCodeBench & \textit{Avg} & MedQA & MMLU-Med & \textit{Avg} & \textit{All} \\
\midrule
No fine-tuning (zero-shot) & 54.54 & 42.91 & 27.50 & 16.42 & 35.34 & 55.08 & 29.04 & 42.06 & 51.40 & 54.84 & 53.12 & 43.51 \\
\midrule
Qwen3-Emb+BM25 & 79.78 & 52.73 & 34.17 & 17.16 & 45.96 & 66.06 & \textbf{32.37} & 49.21 & 49.36 & 58.82 & 54.09 & 49.75 \\
monoT5 (OOD)+BM25 & 69.34 & 48.87 & 29.17 & 17.15 & 41.13 & 57.93 & 31.78 & 44.85 & 52.58 & \textbf{63.66} & 58.12 & 48.03 \\
Claude Code & 77.05 & 45.25 & 30.00 & 17.16 & 42.37 & 64.02 & 29.97 & 47.00 & \textbf{55.77} & 61.25 & 58.51 & 49.29 \\
Codex & 65.56 & 50.82 & 30.83 & 17.16 & 41.09 & 62.60 & 29.73 & 46.17 & 53.44 & 61.86 & 57.65 & 48.30 \\
\midrule
AgenticDS (ours) & \textbf{82.66} & \textbf{57.01} & \textbf{40.00} & \textbf{19.61} & \textbf{49.82} & \textbf{67.07} & 32.16 & \textbf{49.62} & 55.72 & 62.28 & \textbf{59.00} & \textbf{52.81} \\
\bottomrule
\end{tabular}
\end{table*}

\section{Results}
\label{sec:results}

\subsection{Dataset Retrieval (RQ1)}
Table~\ref{tab:retrieval} reports all 15 systems. AgenticDS (Qwen3-Emb) leads on every NDCG cutoff, beating the best agent baseline, Claude Code, by 18.9\% and the strongest hybrid baseline, Qwen3-Emb+BM25, by 25.7\% on NDCG@10. These NDCG margins hold against all fourteen other systems after Holm correction, the weakest of them being the 0.063 gap to our own bge-m3 variant ($p=0.003$). The picture is narrower on MAP: AgenticDS posts the best MAP@10 and MAP@20, but its margins there over the two agent baselines do not survive correction, so we claim an advantage in graded ranking rather than in binary relevance. The same caution applies between our two variants, whose MAP differences are all within noise. Platform search is worst by a wide margin, confirming that literal string matching over inconsistent cards is inadequate for ML training data needs. \emph{AgenticDS outranks all thirteen other systems at every NDCG cutoff, by 18.9\% over the strongest LLM agent and 25.7\% over the strongest hybrid, and the advantage lies in ordering quality, which is what matters when only the top few datasets reach training.}

\textbf{Where does agentic search win most?} The gain is mechanism-aligned rather than uniform: over the hybrid baseline AgenticDS gains $+$0.075 and $+$0.105 NDCG@10 on \emph{general} and \emph{fine-tuning} queries, but $+$0.206 and $+$0.197 on \emph{paper-reference} and \emph{entity-neighbor} queries, so the advantage roughly doubles once an answer stops being recoverable from card text alone. All four gains survive Holm correction over the 56 per-type comparisons (Appendix~\ref{app:extended}). Against the two agent baselines, however, gains separate only on general and fine-tuning queries. On the 40 paper-reference and 30 entity-neighbor queries the differences do not reach the threshold, which we read as insufficient power on those subsets rather than as equivalence. One query makes the mechanism concrete. Asked for ``datasets used to train or finetune the PicoLM-0.5M model'', text retrieval returns benchmarks that merely share the model's name, whereas the Planner makes the training-source relation a hard constraint and the Engineer reaches three supported corpora whose cards never mention the model at all. \emph{The gain therefore tracks the mechanism that produces it, widening where an answer has to be reached through a relation}, which is the case the typed anchor constraint and the whitelisted traversal were designed for.

\subsection{Component Ablation}
Figure~\ref{fig:abl_retr} ablates the AgenticDS search pipeline on HuDS-Bench over ten configurations. This is our primary ablation: it is measured on all 200 queries, every configuration loses accuracy relative to the full system, and the three largest losses all involve the Critic. The Critic and the Planner dominate ($-$0.149 and $-$0.086 NDCG@10), KG expansion follows at $-$0.084, and the remaining three components fall between $-$0.017 and $-$0.036. The feedback-loop entry removes replanning entirely, stopping after the first round, so its $-$0.036 measures what the whole revision mechanism buys. The adaptive budget is the Engineer's per-branch allocation of the anchor, neighbor, and pool budgets, replaced here by an equal split, and sample evidence is the sample fields in the Critic's evidence package, withheld here. \emph{Verification is therefore the load-bearing component, since every configuration that removes the Critic, alone or in combination, falls below both agent baselines.}

One combination is not additive. Removing the Planner \emph{and} the Critic ($-$0.121) is less damaging than removing the Critic alone ($-$0.149), because without a Planner the Engineer falls back to a wide default retrieval that leaves the unverified ranking more usable candidates. Without the Critic, KG candidates introduce noise and degrade sparse, dense, and hybrid retrieval alike (Appendix~\ref{app:extended}), confirming that relational expansion requires semantic verification. \emph{The components therefore interact rather than contribute independently, so the loop has to be read as a whole even though the Critic carries by far the largest single effect.}

\subsection{Training Utility (RQ2)}
Table~\ref{tab:e2e} answers whether retrieval gains transfer to model gains. AgenticDS leads the other four acquisition methods in all three domains, with a grand mean of 52.81 against 49.75 for the second-best, a $+$3.06 margin. Margins in this subsection are means of the three domain averages, each of which first averages the three base models. The equal-weight 24-block mean gives the same ordering. The advantage over each of the four methods is statistically significant after Holm correction ($p=0.016$ to $0.020$), with AgenticDS ahead in 7 of the nine (domain, base model) blocks against Claude Code, 8 against the hybrid of Qwen3-Embedding-0.6B and BM25, and all nine against Codex and monoT5-based reranking. Under the broader correction that also tests the baselines against each other, the margins over Codex and monoT5 remain significant while those over Claude Code and the hybrid do not, and no two baselines separate from one another. \emph{Ranking quality therefore does convert into model quality, which is the link the search-to-train protocol was built to test and the one that graded judgments alone cannot establish.}

\textbf{Domain-level analysis.} Math shows the clearest gain ($+$3.86 over the next best), driven by the competition benchmarks, where AgenticDS retrieves specialized SFT corpora rather than generic math data. Across base models the pattern is monotone in model weakness: Llama-3.1-8B gains $+$25.12 with AgenticDS against $+$21.81 for the best alternative, Qwen2.5-7B $+$14.66 against $+$7.39, and the strongest base, Olmo-3-7B-Instruct, only $+$3.67 against $+$2.65.

Code and medicine show narrower three-model margins ($+$0.41 and $+$0.49), and in both the aggregate hides the structure. In code, dataset cards are self-documenting enough that dense retrieval is already near-optimal on the strong bases, and AgenticDS separates clearly only on Llama-3.1-8B. In medicine the margin again concentrates on Llama-3.1-8B, where AgenticDS reaches 61.16 against 48.41 for the hybrid baseline, a gap of nearly 13 points on a model whose medical cards are too terse for dense retrieval to exploit. The reranking baseline is not subject to the same failure and reaches 60.31, so what separates methods in medicine is whether the acquisition stage can judge a terse card at all, not whether it uses a graph. Across the nine (domain, base model) blocks AgenticDS wins seven and loses two to Claude Code, both in medicine and both by under one point. \emph{The practically relevant regularity is therefore that the weaker the base model, the more acquisition quality decides the outcome, so the systems that most need good data are exactly the ones an acquisition system helps most.}

\subsection{Downstream Component Ablation}
\begin{table}[t]
\centering
\caption{Search-end and training-end ablation compared. Training entries are means over the nine (domain, base model) blocks, and s.e. is the standard error of that mean.}
\label{tab:abl_train}
\small
\begin{tabular}{lrrrc}
\toprule
& Search & \multicolumn{2}{c}{Training} & Sign \\
\cmidrule(lr){2-2}\cmidrule(lr){3-4}\cmidrule(lr){5-5}
Component removed & $\Delta$NDCG@10 & mean & s.e. & agrees \\
\midrule
Critic & $-$0.149 & $-$1.85 & 1.36 & yes \\
Planner & $-$0.086 & $-$0.60 & 0.92 & yes \\
KG expansion & $-$0.084 & $+$0.10 & 0.56 & no \\
Feedback loop & $-$0.036 & --- & --- & --- \\
Adaptive budget & $-$0.031 & $+$0.43 & 0.85 & no \\
Sample evidence & $-$0.017 & $-$1.43 & 1.15 & yes \\
\bottomrule
\end{tabular}
\end{table}

\begin{figure*}[t]
\centering
\includegraphics[width=\textwidth]{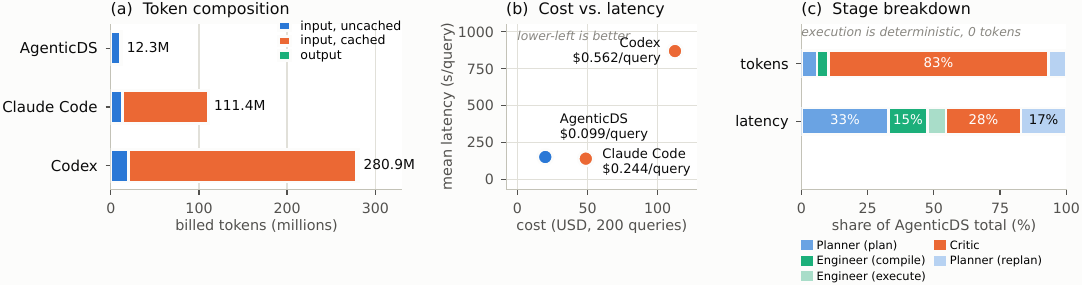}
\caption{Cost over the 200 HuDS-Bench queries. (a) billed tokens by type, (b) cost in USD against mean latency, and (c) AgenticDS token and latency share by stage.}
\Description{Three panels. Left: horizontal stacked bars of billed tokens for AgenticDS (12.3M), Claude Code (111.4M, 87 percent cached) and Codex (280.9M, 92 percent cached). Middle: a scatter of cost in US dollars against mean seconds per query, with AgenticDS cheapest at 19.86 dollars and Codex both slowest and most expensive at 112.40 dollars. Right: two stacked strips over the same five pipeline stages, one for token share and one for latency share, showing the Critic at 83 percent of tokens but 28 percent of time, the two Planner calls at 13 percent of tokens but 50 percent of time, and the Engineer's deterministic execution at 0 percent of tokens and 7 percent of time.}
\label{fig:efficiency}
\end{figure*}

Does the search-end ablation survive fine-tuning? Table~\ref{tab:abl_train} repeats each single-module ablation through the search-to-train protocol, aggregated over all nine (domain, base model) blocks. It does not resolve the question for any individual component. The per-block changes are dispersed enough that a nine-block mean carries a standard error of 0.6--1.4 accuracy points, so a component would have to move that mean by roughly two points to be separated, and the largest effect, removing the Critic at $-$1.85, falls just under the bar. We therefore read no per-component downstream attribution out of this experiment in either direction. In particular, the small positive means for KG expansion and the adaptive budget are not evidence against those modules.

What the experiment does support is a sign check. Three of the five components move both ends in the same direction, and the largest search-end effect is also the most negative downstream. The exception is informative. Sample evidence is the smallest search-end effect ($-$0.017) and the second largest downstream one ($-$1.43), consistent with sample snippets mattering for whether a corpus trains well rather than for whether it ranks well. Appendix~\ref{app:ablation_grid} gives the per-block grid and discusses why a module can be clearly effective at the search stage and invisible after fine-tuning. \emph{The protocol is therefore the right instrument for ranking acquisition methods and the wrong one for assigning credit inside a pipeline, which is why we report its standard error beside its means and keep component attribution at the search end.}

\subsection{Cost Analysis}
Figure~\ref{fig:efficiency} compares AgenticDS with the two agent baselines over the 200 queries, with AgenticDS running under its full $T=3$ cap. AgenticDS matches Claude Code's wall-clock time (150.5 vs.\ 139.3 seconds per query) while billing 12.3M tokens against 111.4M and 280.9M. Most of that token gap is cached prompt reuse billed at a lower rate, so AgenticDS's relative cost savings are smaller than its relative token savings. Because all three systems run the same GLM-5.2 backbone, one price list settles the comparison: the 200-query evaluation costs \$19.86 for AgenticDS against \$48.78 and \$112.40, savings of 59\% and 82\%, and AgenticDS is $5.8\times$ faster than Codex end to end. The saving is a lower bound, since AgenticDS has no cache-hit breakdown and is billed entirely at the uncached input rate. The architectural reason is that the Engineer reads each dataset card at most once and the Critic receives a fixed-format evidence package, whereas free-form tool-calling agents re-fetch and re-read cards during open-ended exploration.

Within AgenticDS the two costs sit in different places. The Critic accounts for 83.0\% of the tokens but only 28\% of the wall clock, the two Planner calls take half the wall clock on 13\% of the tokens, and the retrieval and fusion stage is deterministic and spends no tokens at all. Appendix~\ref{app:implementation} gives the token totals, the per-system costs, and the price list behind these figures. \emph{Bounded autonomy therefore has a measurable payoff. Even with three rounds available, fixing the operator vocabulary and capping the budget holds cost near-constant per query, whereas an open-ended agent's cost grows with however long it chooses to explore.}

\section{Discussion}

\textbf{From retrieval to acquisition.} AgenticDS makes dataset search behave more like product search or paper discovery. It interprets a natural language intent, orchestrates structured and unstructured evidence, and returns candidates that serve the operational purpose of training rather than merely matching a description. Applying the separation between interpretation, execution, and verification that is familiar across Web information systems to ML data acquisition buys a guarantee unconstrained tool-calling agents cannot offer, namely that the same plan reaches the same corpus and yields the same candidate set.

\textbf{The role of KG evidence.} HuggingKG is infrastructure rather than a contribution, and it does two jobs. During search, bounded paths supplement lexical retrieval and record why a resource was reached. During benchmark construction, relation-grounded queries create needs that copying card text cannot meet. The boundary condition on the first job is sharp. Graph candidates added without semantic verification lower NDCG@10 by 0.025--0.029 across three backbones, so expansion and verification are not separable features. Where the graph does pay, it pays as designed: entity-neighbor queries needing one- or two-hop traversal gain $+$0.197 NDCG@10, against $+$0.075 for general queries already answerable from card text. Appendix~\ref{app:cases} traces both behaviours on individual queries, including one the system correctly reports as unsatisfiable.

\section{Conclusion}

In this paper we presented AgenticDS, a contract-governed agentic framework for natural language instructed dataset acquisition on the Web, in which a Planner, an Engineer, and a Critic exchange typed plans, evidence records, and diagnoses so that semantic reasoning never silently rewrites what the retrieval stack does. Experiments on HuDS-Bench, a new benchmark of 200 queries and 24{,}908 graded judgments pooled over 15 systems, show consistent improvements over both commercial LLM agents and strong dense--sparse retrieval pipelines, and a controlled search-to-train study confirms that the datasets AgenticDS returns yield the best downstream accuracy in all three domains we study. Further analysis shows that relational evidence is conditional rather than additive, since graph expansion pays off only once a provenance-aware verifier checks the path behind each candidate. Dataset acquisition can therefore be treated as an upstream stage of a data-centric improvement loop whose quality is judged by the model it trains rather than by relevance alone.

Two limitations bound these claims, since benchmark construction shares a graph with the retrieval engine and the utility study is narrow in coverage. Future work addresses both with a KG-free gold-label pipeline and a distilled Critic. Appendix~\ref{app:limitations} gives the details.

\begin{acks}
This work was supported by the Noncommunicable Chronic Diseases-National Science and Technology Major Project (2025ZD0544900) and the Fundamental Research Funds for the Central Universities (No. 2026300429).
\end{acks}

\bibliographystyle{ACM-Reference-Format}
\bibliography{main}

\appendix
\section{Updated HuggingKG Snapshot}
\label{app:resource}

The experiments use a March 12, 2026 snapshot of HuggingKG~\citep{chen2025benchmarking}. Crawling and normalization yield 4{,}928{,}681 entities (910{,}119 datasets, 2{,}704{,}939 models, 56 tasks, 33{,}524 papers, 1{,}099{,}292 spaces, and 180{,}751 collections) and 3{,}951{,}507 relations. The update changes coverage, not the schema or method. KG expansion resolves typed paper, model, collection, and dataset anchors and traverses one- or two-hop paths under per-branch budgets.

\section{Contracts, Parameters, and Prompts}
\label{app:method}

Contracts specify retrieval hypotheses and query-linked constraints and restrict Engineer execution to admissible operators, declared paths, and configurable query, anchor, neighbor, and pool budgets. Search uses a configurable iteration limit, set to $T=3$ unless stated otherwise, with scoring coefficients and acceptance thresholds fixed across rounds. With preferences, pre-ranking mixes grouped retrieval support and utility through $\beta$ and $1-\beta$. The final score combines semantic judgment and pre-ranking with weights $w_{\rm llm}$ and $1-w_{\rm llm}$. Missing constraint fields remain unknown, and candidates with explicit hard-constraint violations are discarded. Compressed metadata, content, and paths support independent Critic scoring. Acceptance tests configured quality and single-candidate joint-intent targets, at least $K$ candidates, and absence of critical diagnoses. Feedback permits bounded local plan repair tied to the original query. Search retains the best historical round using quality--coverage utility, with configurable weight $\lambda$. The Planner contract, Engineer action protocol, evidence record, and diagnosis fields are summarized in Table~\ref{tab:schema_app}. The full prompts are included in the released code.

\begin{table}[t]
\centering
\caption{Typed objects exchanged between the three agents.}
\label{tab:schema_app}
\scriptsize
\begin{tabular}{p{0.20\columnwidth}p{0.70\columnwidth}}
\toprule
Object & Fields and validation \\
\midrule
Plan $\mathbf P_t$ & Complementary queries; declared typed paths; query-linked hard/soft requirements; intent and acceptance targets. \\
Protocol $\mathbf S_t$ & Ordered predefined operators; index bindings; query, anchor, neighbor, and pool budget limits; all arguments logged. \\
Evidence & Metadata; content snippet; branch ranks; three-valued constraints; typed anchors and paths; field-aware truncation. \\
Diagnosis $\mathcal F_t$ & Failure code/severity; affected intents; evidence and confidence; permitted local plan repair within protocol limits. \\
\bottomrule
\end{tabular}
\end{table}

\section{HuDS-Bench Construction Details}
\label{app:benchmark}

The construction pipeline derives structured conditions (task, modality, language, scale, format) from each gold dataset's textual metadata and KG neighborhood, generates a need-centric query, and verifies that the result does not leak the repository identifier or collapse to a single lookup. Candidate pools merge the top-20 results of all 15 evaluated systems. Each candidate's evidence package contains concise card excerpts, normalized metadata, sample fields and instruction--response snippets when available, and retrieval and graph provenance. Using a structured rubric (released with the code), the LLM annotator marks every constraint \texttt{satisfied}, \texttt{violated}, or \texttt{unknown} and cites an evidence code for each \texttt{satisfied} or \texttt{violated} decision, which we deterministically project to a graded label. Label~2 requires an exact or strong intent match with satisfied hard constraints, a matched dataset role, and direct evidence, label~1 covers partial matches, and label~0 marks hard conflicts or role mismatches. A verifier model re-checks any item with an evidence conflict, a role mismatch, an \texttt{unknown} on a hard constraint, or an unresolved relation, and disputes are escalated to human review. Table~\ref{tab:labels_app} gives the distribution.

\begin{table}[t]
\centering
\caption{Graded relevance judgments by query type.}
\label{tab:labels_app}
\scriptsize
\begin{tabular}{lrrrr}
\toprule
Query type & Pairs & Label 0 & Label 1 & Label 2 \\
\midrule
General & 10{,}730 & 2{,}990 & 6{,}388 & 1{,}352 \\
Fine-tuning & 4{,}968 & 1{,}134 & 3{,}210 & 624 \\
Paper reference & 5{,}049 & 4{,}659 & 1 & 389 \\
Entity neighbor & 4{,}161 & 3{,}842 & 164 & 155 \\
\midrule
Total & 24{,}908 & 12{,}625 & 9{,}763 & 2{,}520 \\
\bottomrule
\end{tabular}
\end{table}

\paragraph{Human agreement study.}
To check that the automated pipeline reproduces human judgment rather than being self-consistent, we drew 2{,}490 distinct query--dataset pairs (10.0\% of the 24{,}908 judged pairs) by random sampling without replacement, spanning 1{,}051 general, 534 fine-tuning, 490 paper-reference, and 415 entity-neighbor pairs, including 861 flagged for review, which were not excluded. One researcher familiar with machine-learning datasets independently applied the same 0/1/2 label criteria, seeing the query, metadata, full dataset card, and the necessary entity-page and graph evidence. Because partial credit is the hardest grade to assign, label~1 is also the grade on which the two relational query types almost never fire (Table~\ref{tab:labels_app}), so the overall agreement score mainly reflects the two semantic types. Comparing the annotator's first labels with the previously frozen automated labels, before any disagreement analysis or correction, yields 84.1\% exact agreement (2{,}095/2{,}490) and linearly weighted Cohen's $\kappa=0.754$, substantial agreement~\citep{landis1977kappa}.

\section{Extended Retrieval Results}
\label{app:extended}

The complete 15-system comparison is reported in the main text (Table~\ref{tab:retrieval}). This appendix reports per-query-type retrieval gains, the agent-backbone sweep, and the Critic-free graph experiment.

\begin{table}[t]
\centering
\caption{NDCG@10 gain over the Qwen3-Emb+BM25 hybrid by query type. All four survive Holm correction.}
\label{tab:qtype}
\scriptsize
\begin{tabular}{llrr}
\toprule
Answer reachable from & Query type & Queries & $\Delta$NDCG@10 \\
\midrule
\multirow{2}{*}{Card text} & General & 90 & $+$0.075 \\
 & Fine-tuning & 40 & $+$0.105 \\
\midrule
\multirow{2}{*}{Relations only} & Paper reference & 40 & $+$0.206 \\
 & Entity neighbor & 30 & $+$0.197 \\
\bottomrule
\end{tabular}
\end{table}

\begin{table}[t]
\centering
\caption{Sensitivity to the agent backbone.}
\label{tab:backbone_app}
\scriptsize
\begin{tabular}{lrrr}
\toprule
Backbone & NDCG@5 & NDCG@10 & MAP@10 \\
\midrule
GLM-5.2 & 0.6297 & 0.6148 & 0.6973 \\
Qwen3.5-35B-A3B & 0.5417 & 0.5244 & 0.5702 \\
GLM-4-9B & 0.4809 & 0.4706 & 0.5750 \\
\bottomrule
\end{tabular}
\end{table}

Table~\ref{tab:qtype} lists the per-type gains summarized in Section~\ref{sec:results}. The two semantic types, which together account for 130 of the 200 queries, gain least. The two relational types, whose gold answers are reachable only through the graph, gain twice as much. The four gains weight-average to the $+$0.126 overall margin of Table~\ref{tab:retrieval}. Because the relational types carry a near-binary label distribution (Table~\ref{tab:labels_app}) and rest on only 40 and 30 queries, their margins are more volatile than the semantic types', so we read the split between the two families rather than the gap between the relational rows.

Table~\ref{tab:backbone_app} also bounds the architecture's claim. It varies only the agent backbone, holding the retrieval stack and all contracts fixed. The contracts constrain \emph{what} the Planner and Critic may emit, not how well they reason: with Qwen3.5-35B-A3B the system still clears every retrieval baseline (0.5244 NDCG@10), but with GLM-4-9B it falls to 0.4706, level with BM25 (0.4717) and below the hybrid baseline (0.4893). The contract-governed loop therefore converts backbone capability into ranking quality without substituting for it, and a frontier-scale Critic is a prerequisite, not an optimization.


\section{Significance Testing}
\label{app:significance}

We follow standard practice for ranking comparisons~\citep{smucker2007comparison}. Retrieval comparisons take the per-query difference against AgenticDS (Qwen3-Emb) and run a two-sided paired permutation test with Holm correction, with paired bootstrap intervals stratified by query type. Training comparisons block by the nine (domain, base model) pairs, average benchmarks within a domain, and enumerate all $2^9$ sign assignments, making that test exact rather than sampled.

\paragraph{Outcomes.}
All fourteen retrieval comparisons are significant on NDCG@5, NDCG@10, and NDCG@20. On MAP the comparisons against Claude Code and Codex are not, with Holm $p$ between 0.13 and 0.45, and neither are the differences between our two variants. Correcting jointly across metrics leaves every NDCG@10 comparison significant except the one against our own bge-m3 variant ($p=0.063$), and the two LLM agents do not separate from each other ($\Delta=0.003$, $p=1.00$). On the training side the four comparisons against AgenticDS give raw $p$ of 0.0195 (Claude Code), 0.0039 (Codex), 0.0078 (hybrid), and 0.0039 (monoT5), all significant after Holm correction, and a Wilcoxon signed-rank test agrees.

\paragraph{Scope.}
The training result is sensitive to the pairing unit, since collapsing to three domain-level units leaves only $2^3$ sign assignments and a smallest two-sided $p$ of 0.25, so we report every grouping rather than the best. More broadly, every system contributes a single run, the 200 queries come from graph anchors and are not fully independent, the graded labels are silver, and the training side has no independent seeds, none of which Holm correction repairs.

\section{Training-Study Details}
\label{app:trainingdetails}

\subsection{Benchmark configuration}

Table~\ref{tab:benchconfig} fixes the evaluation split and metric of every benchmark so that absolute numbers are comparable with other work. All scores are normalized to the 0--100 range.

We evaluate the 272-problem OCWCourses set of~\citep{lewkowycz2022minerva}, not the 5{,}000-problem MATH split that some harnesses report under the same name. \emph{MMLU-Medical} is the union of nine subjects from the \texttt{cais/mmlu} test splits, used as published: anatomy, clinical knowledge, college biology, college medicine, human aging, medical genetics, nutrition, professional medicine, and virology (1{,}784 items).

\begin{table}[t]
\centering
\caption{Evaluation split of every benchmark.}
\label{tab:benchconfig}
\scriptsize
\begin{tabular}{llrl}
\toprule
Domain & Benchmark & Items & Metric \\
\midrule
\multirow{4}{*}{Math} & GSM8K & 1{,}319 & exact match \\
& MATH & 5{,}000 & exact match \\
& AMC23 & 40 & exact match \\
& OCWCourses & 272 & exact match \\
\midrule
\multirow{2}{*}{Code} & HumanEval & 164 & pass@1 \\
& BigCodeBench & 1{,}140 & pass@1 \\
\midrule
\multirow{2}{*}{Medicine} & MedQA & 1{,}273 & accuracy \\
& MMLU-Medical & 1{,}784 & accuracy \\
\bottomrule
\end{tabular}
\end{table}

\begin{figure*}[t]
\centering
\includegraphics[width=0.8\textwidth]{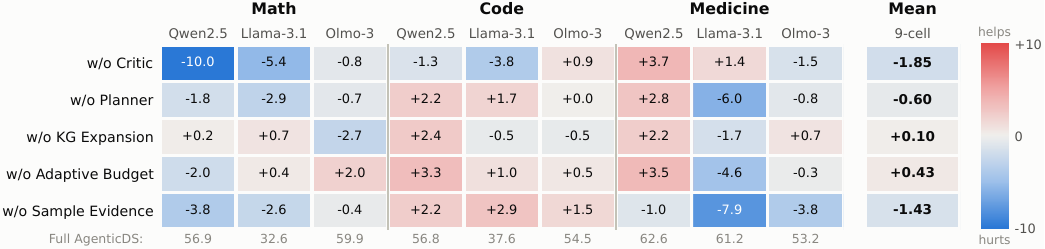}
\caption{Component ablation under the search-to-train protocol, all base models and domains. Each block is the change in domain accuracy when a component is removed. The full system's score is printed beneath each column.}
\Description{Heatmap of five ablation rows by nine base-model-and-domain columns, coloured on a blue-to-red diverging scale centred at zero, with the numeric delta printed in every block and a separated nine-block mean column at the right.}
\label{fig:abl_grid}
\end{figure*}

\subsection{Training grid and search queries}

For training interventions, every method uses the same eligibility rules, $b=2000$ per dataset, an identical optimizer, batch size = 4, and a grid of three epochs and learning rates in $\{10^{-5},5{\times}10^{-6},10^{-6}\}$ chosen on a held-out development split. 

The \emph{math} query reads ``advanced competition mathematics instruction-tuning data with difficult problems, complete derivations, and verified boxed answers.'' The \emph{code} query reads ``competitive programming instruction-tuning data with problem statements, test cases, accepted code solutions, and code generation supervision.'' The \emph{medicine} query reads ``medical question-answering SFT data with professional multiple-choice questions, clinical explanations, and final answers across broad specialties.''

\subsection{Per-model scores}

In the training study, AgenticDS denotes the Qwen3-Emb variant. Table~\ref{tab:e2e_full_app} expands the means of Table~\ref{tab:e2e} into their nine blocks.

\begin{table}[t]
\centering
\caption{Per-model domain scores. Bold: best per row.}
\label{tab:e2e_full_app}
\resizebox{\columnwidth}{!}{\scriptsize
\begin{tabular}{llrrrrr}
\toprule
Domain & Model & Qwen3-Emb+BM25 & monoT5 (OOD) & Claude Code & Codex & AgenticDS \\
\midrule
\multirow{3}{*}{Math} & Qwen & 49.63 & 46.40 & 39.60 & 45.84 & \textbf{56.90} \\
& Llama & 29.34 & 21.79 & 29.20 & 19.45 & \textbf{32.64} \\
& Olmo & 58.90 & 55.21 & 58.30 & 57.98 & \textbf{59.93} \\
\midrule
\multirow{3}{*}{Code} & Qwen & 56.58 & 52.79 & 55.48 & 53.56 & \textbf{56.80} \\
& Llama & 36.86 & 36.69 & 34.32 & 32.23 & \textbf{37.55} \\
& Olmo & 54.20 & 45.09 & 51.19 & 52.71 & \textbf{54.50} \\
\midrule
\multirow{3}{*}{Medicine} & Qwen & 62.59 & 61.54 & \textbf{63.47} & 62.17 & 62.58 \\
& Llama & 48.41 & 60.31 & 57.99 & 59.28 & \textbf{61.16} \\
& Olmo & 51.28 & 52.51 & \textbf{54.08} & 51.52 & 53.25 \\
\bottomrule
\end{tabular}}
\end{table}

\subsection{Ablation grid}
\label{app:ablation_grid}

Figure~\ref{fig:abl_grid} gives the full three-base $\times$ three-domain $\times$ five-ablation grid, computed under the eight-benchmark grid of Section~\ref{sec:evaluation} with the per-benchmark best learning rate, then averaged within a domain. Table~\ref{tab:abl_train} aggregates these 45 blocks into the five row means and standard errors discussed in Section~\ref{sec:results}.

Blue blocks mean that removing the component costs accuracy and red blocks that removing it gains accuracy. Removal helps in 21 of the 45 blocks, and the per-block changes scatter widely enough (standard deviation 1.7--4.1 points) that the standard error of any nine-block mean is 0.6--1.4 points. No component clears two standard errors, so the means for KG expansion ($+0.10$) and the adaptive budget ($+0.43$) should be read as \emph{unresolved} rather than as evidence against those modules. The useful content of this figure is the dispersion itself, which is why Section~\ref{sec:results} reports the aggregate with its standard error rather than per-block wins.

\section{Implementation and Cost Comparison}
\label{app:implementation}

\paragraph{Full baseline inventory.}
Beyond the four strong baselines detailed in Section~\ref{sec:evaluation}, nine more span four families.
\begin{itemize}
  \item \emph{Platform search}: Google Dataset Search~\citep{brickley2019google} and HuggingFace full-text search.\footnote{\url{https://huggingface.co/search/full-text}}
  \item \emph{Sparse}: BM25~\citep{robertson2009bm25} and BM25+RM3~\citep{abduljaleel2004rm3}.
  \item \emph{Dense}: bge-m3~\citep{chen2024m3} and Qwen3-Embedding-0.6B~\citep{zhang2025qwen3}.
  \item \emph{Hybrid}: bge-m3+BM25, MDR-BM25~\citep{chen2020schema}, and ANCE+BM25~\citep{xiong2021ance}.
\end{itemize}
All hybrids fuse their two branches with RRF~\citep{cormack2009rrf} under the same smoothing constant, and monoT5 and ANCE are marked \emph{(OOD)} because both are applied zero-shot.

\begin{table}[t]
\centering
\caption{End-to-end cost of the 200-query evaluation. A dash marks no cached-prompt reuse, so AgenticDS is billed entirely at the uncached rate and its cost is an upper bound.}
\label{tab:cost_compare}
\scriptsize
\setlength{\tabcolsep}{3.2pt}
\begin{tabular}{lrrrrrr}
\toprule
& & \multicolumn{3}{c}{Billed tokens (M)} & \multicolumn{2}{c}{Cost (USD)} \\
\cmidrule(lr){3-5}\cmidrule(lr){6-7}
System & s/query & in, unc. & in, cached & output & total & per query \\
\midrule
AgenticDS (Qwen3-Emb) & 150.5 & 10.6 & --- & 1.7 & 19.86 & 0.099 \\
AgenticDS (bge-m3) & 212.7 & 13.1 & --- & 2.1 & 24.35 & 0.122 \\
Claude Code & 139.3 & 13.1 & 97.2 & 1.0 & 48.78 & 0.244 \\
Codex & 868.5 & 20.0 & 258.1 & 2.8 & 112.40 & 0.562 \\
\bottomrule
\end{tabular}
\end{table}

Table~\ref{tab:cost_compare} prices every system against the same GLM-5.2 list, \$1.19 per million uncached input tokens, \$0.30 per million cached input tokens, and \$4.17 per million output tokens.

Dense search uses separate FAISS indexes for descriptions and concatenated metadata, while sparse search uses a Lucene BM25 index. All hyperparameters (RRF smoothing, dense/sparse fusion weight, $w_{\rm llm}$, acceptance thresholds) were tuned on an internal pilot set drawn \emph{before} HuDS-Bench was released, and no query from the 200-query release is used for tuning.

\paragraph{LLM agent baseline configuration.}
We implement the two LLM agents with the Claude Code~2.1.228 and Codex~0.147.0 harnesses, both using GLM-5.2 and the same HuggingFace hub tools.

Each agent was invoked with (1)~the raw HuDS-Bench query as the user turn, (2)~a system prompt requesting 30 unique, ranked HuggingFace dataset identifiers satisfying the query, (3)~shared read-only MCP tools for dataset search and metadata/card inspection, and (4)~a JSON output schema for the ranked identifiers. Outputs are normalized, deduplicated, and filtered once against the evaluation snapshot to retain up to 20 datasets, and short lists remain unfilled. Each query was run once, with retries only upon execution errors, so the comparison is single-shot. System prompts, tool specifications, and raw response logs are released with HuDS-Bench.

\section{Case Studies}
\label{app:cases}

Two cases show how contracts organize evidence.
(1)~For the fine-tuning query ``\emph{instruction tuning data with code commit messages as natural language instructions across programming languages}'', the Planner separates code generation, supervision format, and commit provenance into complementary hypotheses. The Engineer's dense branch returns four commit-oriented corpora in the top-10, two of which the Critic rejects because their sample fields show synthetic rather than real-commit prompts, and those rejections guide a local repair hypothesis. (2)~For the general query ``\emph{large Hungarian-only medical consultation data}'', none of the top five results is relevant. The Critic diagnoses unmet language and domain requirements, and local repair must preserve the Hungarian-only constraint, which broader Finno-Ugric or English candidates cannot satisfy. Reporting an unresolvable constraint is the correct behavior here, and separates AgenticDS from both commercial agent baselines, which silently returned near-miss English or Spanish corpora.

\section{Limitations and Future Work}
\label{app:limitations}

\textbf{Limitations.} (1)~The two relational query types are built from HuggingKG anchors that the Engineer can also traverse. The design bounds what that overlap can buy: 130 of the 200 queries carry no KG-derived gold label, every non-KG baseline is given HuggingFace's full-text search, and the variant without KG expansion still reaches 0.5307 NDCG@10, above every non-agent baseline. Extending construction to KG-free gold labels would remove the overlap entirely. (2)~With one query per domain across three base models, the utility study is powered to separate acquisition methods in aggregate rather than individual components. Its scope is English queries on a single ML hub, at a latency suited to offline acquisition rather than interactive search. (3)~The agreement study reports a single $\kappa$ over all four query types. Because the relational types rarely use the middle grade, that figure speaks mainly for the semantic types, and per-type coefficients would be more informative.

\textbf{Future work.} Four directions follow. Sharpening the search-to-train protocol calls for more queries per domain and repeated seeds rather than more benchmarks. A gold-label pipeline that never consults the graph would settle the overlap above. Distilling the Critic into a smaller pairwise scorer attacks the dominant cost. Finally, only the Engineer is tied to a hub, so any ecosystem with typed metadata and a navigable relation graph could be served by swapping it. A federation layer routing typed constraints to several hubs would test the contracts outside HuggingFace.

\end{document}